\pdfoutput=1

\documentclass[11pt]{article}

\usepackage[]{acl}

\usepackage{times}
\usepackage{latexsym}

\usepackage{graphicx}
\usepackage{multirow}
\usepackage{array}
\usepackage{booktabs}

\usepackage[T1]{fontenc}

\usepackage[utf8]{inputenc}

\usepackage{microtype}

\title{Improving Multi-Party Dialogue Discourse Parsing \\via Domain Integration}

\author{Zhengyuan Liu, \ Nancy F. Chen \\
  Institute for Infocomm Research, A*STAR, Singapore \\
  \texttt{\{liu\_zhengyuan,nfychen\}@i2r.a-star.edu.sg}}

\date{}

\begin{document}
\maketitle
\begin{abstract}
While multi-party conversations are often less structured than monologues and documents, they are implicitly organized by semantic level correlations across the interactive turns, and dialogue discourse analysis can be applied to predict the dependency structure and relations between the elementary discourse units, and provide feature-rich structural information for downstream tasks. However, the existing corpora with dialogue discourse annotation are collected from specific domains with limited sample sizes, rendering the performance of data-driven approaches poor on incoming dialogues without any domain adaptation. 
In this paper, we first introduce a Transformer-based parser, and assess its cross-domain performance. We next adopt three methods to gain domain integration from both data and language modeling perspectives to improve the generalization capability. Empirical results show that the neural parser can benefit from our proposed methods, and performs better on cross-domain dialogue samples.
\end{abstract}

\section{Introduction}
\label{introduction}
Text-level discourse parsing is to convert a piece of text into a structured format, by identifying the links and relations between Elementary Discourse Units (EDUs). Incorporating discourse information is proved beneficial for various natural language processing tasks such as machine comprehension \cite{narasimhan-barzilay-2015-machine} and summarization \cite{xu-etal-2020-discoBERT}. Since discourse parsing is involved in capturing and comprehending various semantic and pragmatic phenomena as well as understanding the structural discourse properties, it is quite challenging for machines to conduct automatic processing. There are a series of studies that provide theories and data for developing computational solutions, such as the Penn Discourse Treebank (PDTB) \cite{prasad2008pennPDTB} with sentence-level annotation, and the Rhetorical Structure Theory (RST) \cite{carlson2002rst} with document-level annotation. In RST treebanks, each processed passage is in a hierarchical constituency-based tree structure, and adjacent EDUs are merged to form larger spans\footnote{The merged spans are named as complex discourse units (CDUs) in which multiple EDUs and/or CDUs are grouped together to form a single argument to a discourse relation \cite{asher2016STAC}} recursively \cite{li2014recursive}.

\begin{figure}[t!]
    \begin{center}
    \includegraphics[width=0.47\textwidth]{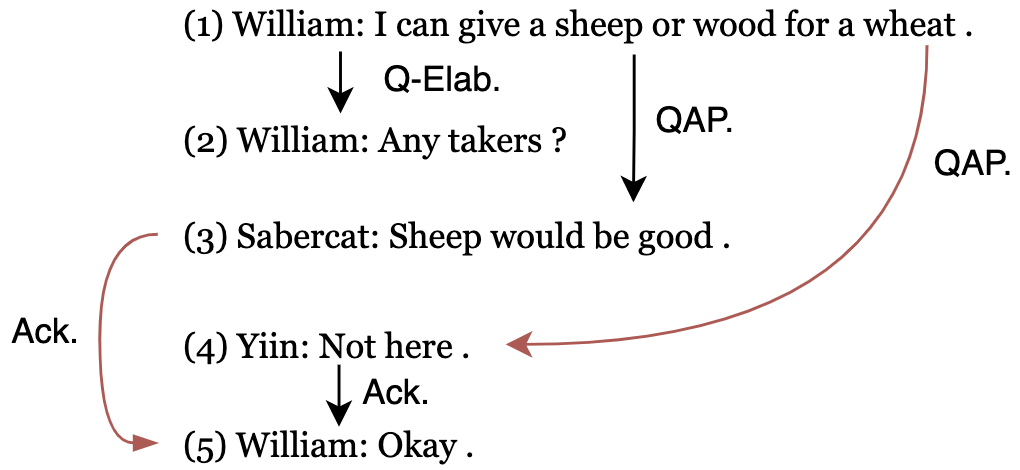}
    \end{center}
    \caption{A multi-party dialogue example \cite{shi2019deepSeqParser} with discourse link and relation annotation in the STAC Corpus \cite{asher2016STAC}. ``Ack.'' is short for relation ``Acknowledgement'', ``QAP.'' for ``Question-Answer-Pair'', and ``Q-Elab.'' for ``Question-Elaboration''. The links in red form a non-projective structure \cite{mcdonald2005non-projective}.}
    \label{fig:stac-example}
\vspace{-0.2cm}
\end{figure}

Recently, the Segmented Discourse Representation Theory (SDRT) is proposed for multi-party dialogue discourse parsing \cite{asher2003logics,asher2016STAC}, which is different from RST whose annotations are on documents. Additionally, SDRT-based annotations contain non-projective links. For example, as shown in Figure \ref{fig:stac-example}, a discourse structure will become non-projective when it is impossible to draw the relations on the same side without crossing \cite{mcdonald2005non-projective}. In this case, the constituency-based structure is not applicable. As a result, the SDRT proposed to transform dialogue discourse trees to a dependency-based structure, where EDUs are directly linked to their precedents without forming upper-level spans.

\begin{figure}[t!]
    \begin{center}
    \includegraphics[width=0.48\textwidth]{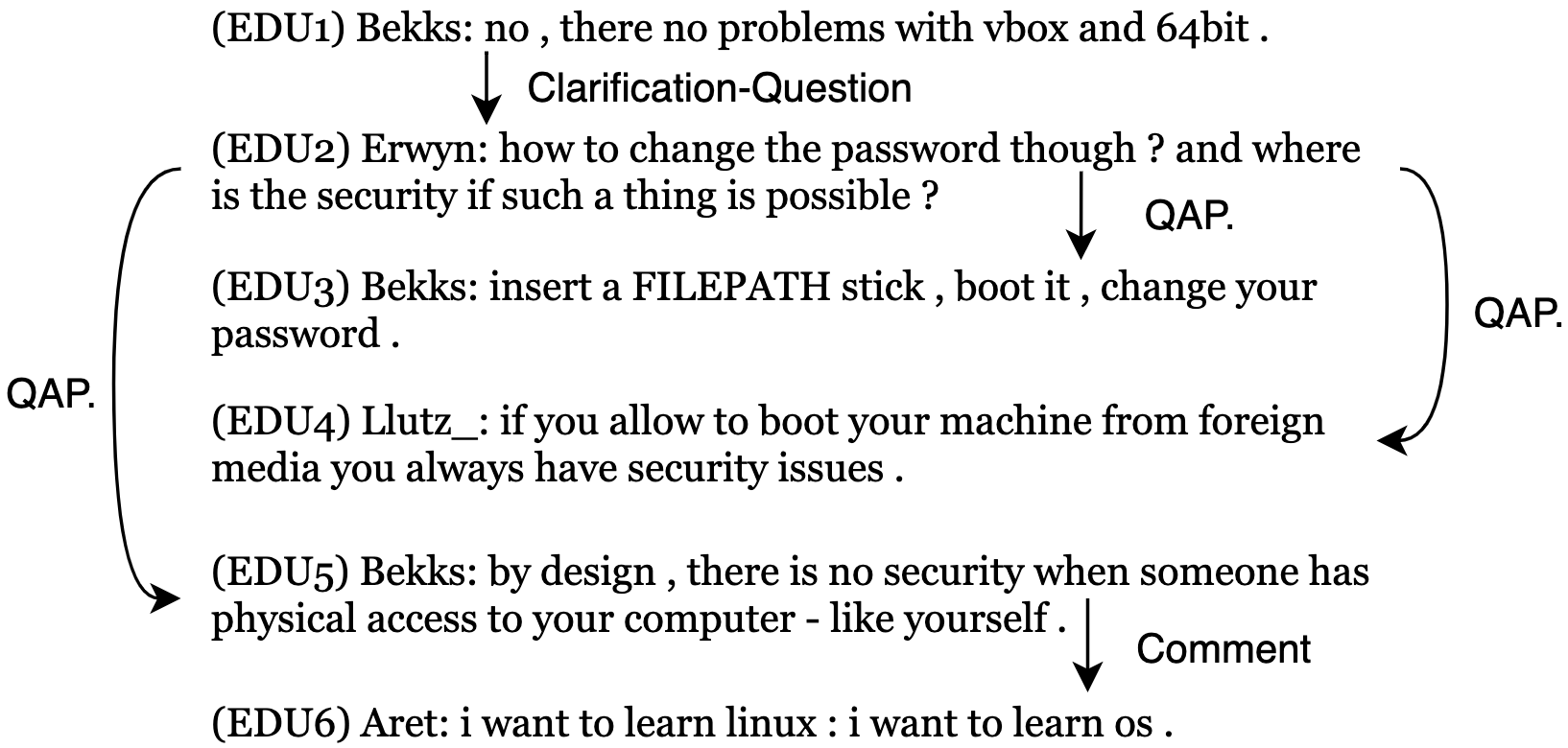}
    \end{center}
     \caption{A multi-party dialogue example with its discourse annotation from the Molweni \cite{li2020molweni}.}
    \label{fig:mol-example}
\end{figure}

\begin{table}[t!]
\linespread{1.0}
    \centering
    \small
    \begin{tabular}{p{2.9cm}ccc}
    \toprule
        & \textbf{RST-DT} & \textbf{STAC} & \textbf{Molweni} \\
    \midrule
         Training Sample Size & 347 & 1091 & 9000 \\
         Test Sample Size & 38 & 100 & 500 \\
         Average EDU Number & 56.03 & 10.95 & 8.82 \\
         Average Word Number & 531.8 & 46.7 & 96.1 \\
         Annotation Scheme & RST & SDRT  & SDRT \\
         Relation Number & 18 & 17 & 17 \\
         Data Domain & News & Game & Ubuntu \\
         Conversational Data & No & Yes & Yes \\
    \bottomrule
    \end{tabular}
    \caption{Data statistics of training samples from three text-level discourse parsing treebanks.}
    \label{tab:dataset}
\vspace{-0.2cm}
\end{table}

Since manual parsing is labor-intensive and time-consuming, automatic discourse analysis under the SDRT theory raises research interest \cite{badene2019weakSupervision}. Previous models show reasonable results on benchmark treebanks \cite{shi2019deepSeqParser}, and utilizing structural information benefits follow-up applications such as dialogue summarization \cite{feng2020-DDAGCN}.
However, domain generality is less studied yet important in practical use cases. Existing treebanks only contain limited training data (as shown in Table \ref{tab:dataset}) and limited domain coverage. An SDRT parser trained on strategic game conversations \cite{asher2016STAC} may not perform well on technical discussions \cite{li2020molweni}, and the sub-optimal parsing could further affect downstream task performance.  Moreover, due to the annotation complexity, the labeled samples from various domains are not readily available for transfer learning \cite{yu2019transfer}.

In this paper, we evaluate and improve the cross-domain generality of neural dialogue discourse parsing:
(1) we conduct a statistical analysis on existing dialogue discourse treebanks, and figure out the possible factors resulting in the gap across multiple domains from a data perspective;
(2) we introduce a Transformer-based neural model for the dependency-based discourse parsing;
(3) we propose three methods for better sharing the effective features across dialogue domains: utilizing prior language knowledge, cross-domain pre-training, and vocabulary refinement. Experimental results on STAC \cite{asher2016STAC} and Molweni \cite{li2020molweni} show that the parsing performance of single-domain training drops significantly on the out-of-domain samples, and it can be improved by our proposed methods.

\begin{figure}[t!]
    \begin{center}
    \includegraphics[width=0.25\textwidth]{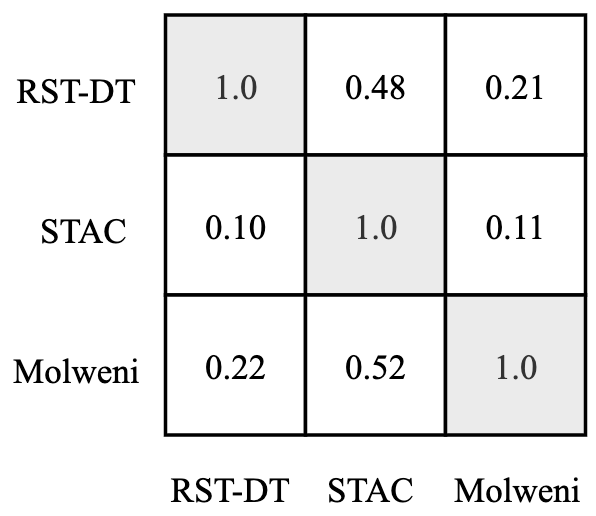}
    \end{center}
     \caption{Word level vocabulary overlap of three text-level  discourse treebanks. The vocabulary sizes of RST-DT, STAC, and Molweni are 17824, 3642, and 18936, respectively.}
    \label{fig:vocab-overlap}
\end{figure}

\begin{figure}[t!]
    \begin{center}
    \includegraphics[width=0.48\textwidth]{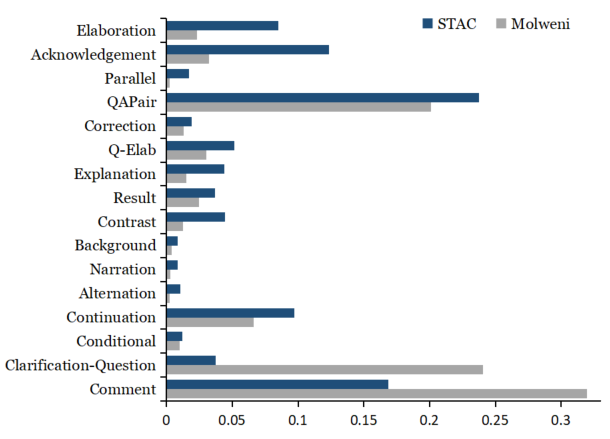}
    \end{center}
    \caption{Discourse relation distributions of STAC and Molweni. X axis denotes the label frequency.}
    \label{fig:relation-dist}
\end{figure}

\begin{figure*}[ht!]
    \begin{center}
    \includegraphics[width=1.0\textwidth]{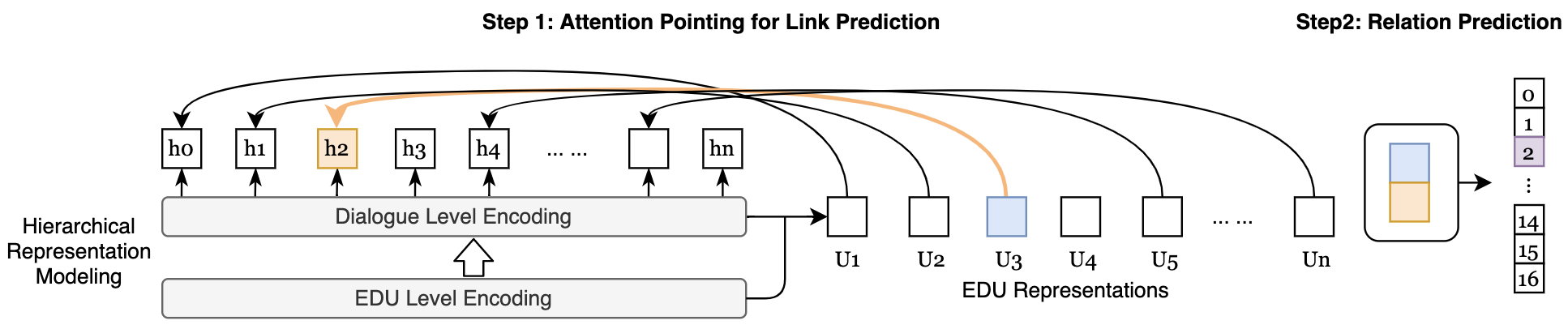}
    \end{center}
    \caption{Overview of the dependency-based discourse parsing framework.}
    \label{fig:framework}
\end{figure*}

\section{Corpora Analysis}
\label{sec:corpus_analysis}
In this section, we conduct a statistical analysis of three text-level discourse treebanks for data-related factors that potentially affect model generality.
\newline\noindent \textbf{RST Discourse Treebank} (RST-DT) \cite{carlson2002rst} is the first corpus for text-level document parsing, and contains articles from the Wall Street Journal (WSJ). While it is not in the dialogue domain, we include it for an extensive comparison.
\newline\noindent \textbf{STAC} \cite{asher2016STAC} is the first corpus for multi-party dialogue discourse parsing, and built on 1.2k strategic conversations where participants take discussion during playing an online game.
\newline\noindent \textbf{Molweni} \cite{li2020molweni} follows the same annotation scheme as STAC, and the data (12k samples) are collected from an online forum, where people discuss technical topics about the Ubuntu system.

The data statistics are summarized in Table \ref{tab:dataset}. (1) Compared with Molweni, the RST-DT and STAC have much smaller sample sizes. (2) Samples from RST-DT have a larger EDU number than STAC and Molweni, resulting in deeper parsed tree structures. The tree depth is one of the major factors that affect parsing complexity. (3) Interestingly, while the word number of Molweni is two times larger than that of STAC, no significant difference in their average EDU numbers, resulting in a similar parsing complexity from a depth perspective. (4) The lexical distributions of STAC and Molweni are significantly different sharing a small portion of common vocabulary (Figure \ref{fig:vocab-overlap}), as they focus on different conversation scenarios (\textit{Game} vs. \textit{Ubuntu}). (5) Despite the domain distinction between STAC and Molweni, their relation distributions are similar, except that frequencies of the relation (\textit{Clarification-Question} and \textit{Comment}) are quite different, probably because the online technical forums contain more question-clarification and comments (Figure \ref{fig:relation-dist}). While STAC and Molweni are annotated under the same SDRT theory, their lexical features and relation distributions are different, which we speculate will influence the domain generality.

\begin{figure}[t!]
    \begin{center}
    \includegraphics[width=0.42\textwidth]{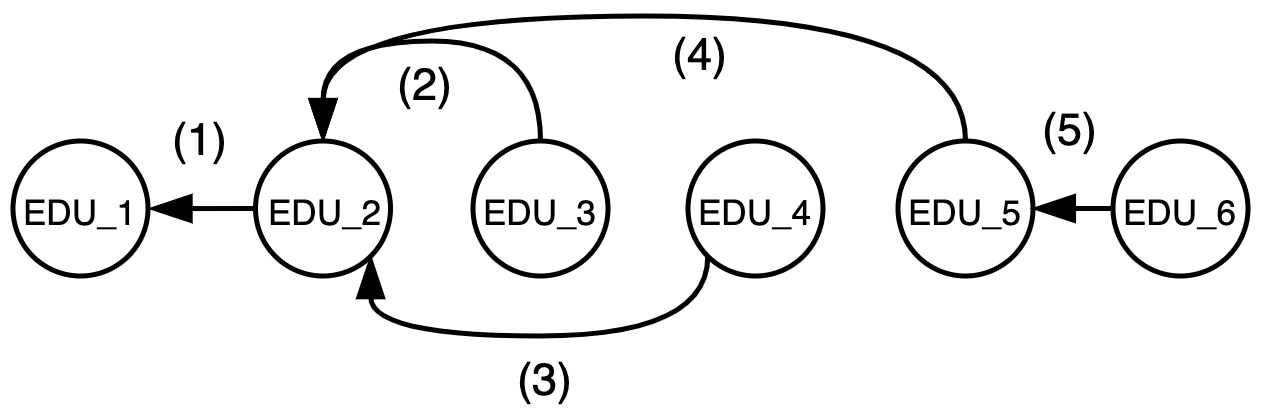}
    \end{center}
     \caption{Illustration of parsing process on the dialogue example shown in Figure \ref{fig:mol-example}. Numbers in brackets denote the order of link prediction, which is in a sequential manner. This produces a dependency structure.}
    \label{fig:link-process}
\vspace{-0.2cm}
\end{figure}

\section{Dialogue Discourse Parsing}
\subsection{Task Definition}
Given a dialogue that has been segmented into a sequence of EDUs $\{u_0, u_1,..., u_n\}$ where $n$ is the EDU number, the discourse parser is applied to predict links and the corresponding relation types between the EDUs. The predicted structure constitutes a dependency tree, which is a special type of Directed Acyclic Graph (DAG). 
As in previous work \cite{shi2019deepSeqParser}, each EDU is only linked to one of their precedent EDUs, and there are no backward links.
As shown in Figure \ref{fig:link-process}, the parsing process can be conducted by a sequential scan of the EDUs.
For one EDU $u_i$, the model predicts a dependency link by estimating a probability distribution as $P(u_j|u_i, U^{pair}_i)$ where $0\leq j<i$ and $U^{pair}_i=\{(u_l,u_k,r_{l,k})|0\leq l<k<i\}$ is the set of already predicted pairs before the current step $i$. The model then determines the relation type based on the predicted link $P(r_{i,j}|u_i, u_j)$ where $j < i$ and $r_{i,j}$ is in the range of $[0,C]$ ($C$ is the number of relation types). Following \citet{li2014textParsing}, we add a \textit{root} node as $u_0$, and if one EDU is not linked from any preceding nodes, it is pointed to $u_0$.

\subsection{Transformer-Based Discourse Parser}
In this paper, based on the sequential parsing process \cite{shi2019deepSeqParser}, we introduce a Transformer-based model for dialogue discourse parsing (as shown in Figure \ref{fig:framework}), which is comprised of the following components:
\newline\noindent \textbf{Hierarchical Encoder.} The encoder computes EDU global representations in a hierarchical manner. A Transformer encoder \cite{vaswani-2017-Transformer} is used for token-level encoding.\footnote{Due to space limitation, refer to \cite{vaswani-2017-Transformer} for more details of the Transformer architecture.}
\begin{equation} H_{token}=\mathrm{TransformerEnc}([t_0, t_1,..., t_m])
\end{equation}
where $t$ denotes token, and $m$ denotes token number. For the $i$-th EDU, its local representation $h^i_{edu}$ is obtained by averaging\footnote{We also adopt \textit{first-and-last} sum and \textit{only-first} sum for EDU representation, and the averaging performs best.} its corresponding tokens hidden states. Then the local EDU representations are fed to a bi-directional GRU component \cite{chung2014GRU} for dialogue-level encoding, and we get final representations $H'$ with both local and global information.
\begin{equation} h'_{i}=[\mathrm{GRU}^{Forward}_{h^i_{edu}};\mathrm{GRU}^{Backward}_{h^i_{edu}}]
\end{equation}

\noindent\textbf{Link Prediction.} An attentive pointer network \cite{vinyals2015pointer} is used for the link prediction. For the $i$-th EDU, we compute a list of attentive scores with a linear layer between the current node and each candidate $h'_i$ where $j<i$. Then scores are normalized by softmax function to a distribution over the previous EDUs, and we obtain the linked EDU with the largest pointing probability.
\begin{equation}
\normalsize
    s_{i,j} = \mathrm{Linear}([h'_i; h'_j])
\end{equation}
\vspace{-0.4cm}
\begin{equation}
\normalsize
    a_{i,j} = \frac{\mathrm{exp}(s_{i,j})}{\sum_{j=0}^i\mathrm{exp}(s_{i,j})}
\end{equation}

\noindent\textbf{Relation Classification.} Given one linked pair is $h'_i$ and $h'_j$, we concatenate and feed them to a relation classifier (a linear component):
\begin{equation}
\normalsize
    r_{i,j} = \mathrm{Linear}([h'_i; h'_j])
\end{equation}
then the output is a probability over the 17 pre-defined discourse relations. For link and relation prediction, the negative log-likelihood is adopted for the loss function.

\section{Cross-Domain Integration}
Based on the corpora analysis in Section \ref{sec:corpus_analysis}, to improve the domain-level generality, we investigate three methods to encourage the neural model to utilize the shared linguistic features from different dialogue domains.
\newline\noindent \textbf{Utilizing Language Backbone.}
Large-scale pre-trained language models provide feature-rich contextualized representations \cite{devlin-2019-BERT}. In previous work, utilizing prior knowledge can boost the performance in parsing tasks, and also shows some but still limited generalization capability at domain and language level \cite{liu2020multiRST}.
Here, we select the \textit{`RoBERTa-base'} model \citep{liu2019roberta} as the language backbone.
\newline\noindent \textbf{Cross-Domain Pre-training.}
Following \citet{dontstoppretraining2020}, we conduct the masked language modeling pre-training with the joint data of STAC and Molweni. This can fuse dialogue-related linguistic features to the language backbone, which is not pre-trained on human conversations. Moreover, pre-training with multiple data resources can increase the domain coverage, and this step (parsing annotation is not required) can be conducted before the task-specified learning.
\newline\noindent \textbf{Cross-Domain Vocabulary Refinement.}
In Section \ref{sec:corpus_analysis}, we observe that the vocabulary overlap between STAC and Molweni is limited (see Figure \ref{fig:vocab-overlap}). Dialogues in Molweni contain a certain amount of technical-related words, whereas STAC contains more game-related words. As the model may overfit corpus-specified lexical features, a vocabulary refinement is adopted by filtering out words that are in lower frequency ($<20$ occurrence) and not shared by the two datasets.

\begin{table}[t!]
\linespread{1.0}
\centering
\small
\begin{tabular}{p{3.2cm}p{1.5cm}<{\centering}p{1.5cm}<{\centering}}
\toprule
\textbf{Train on Joint Data}  & \textbf{Link} & \textbf{Link+Rel.} \\
\midrule
\multicolumn{3}{l}{Deep Sequential Parser \cite{shi2019deepSeqParser}} \\
Test on STAC & 72.8 & 54.8 \\
Test on Molweni & 77.4 & 54.3 \\
\midrule
\multicolumn{3}{l}{Our Proposed Parser w/ language backbone} \\
Test on STAC & \textbf{75.5} & \textbf{57.2} \\
Test on Molweni & \textbf{80.2} & \textbf{56.9} \\
\bottomrule
\end{tabular}
\caption{\label{result-join-table} F1 scores of link and relation prediction with models trained on the joint data (\textbf{STAC+Molweni}).}
\end{table}

\begin{table}[t!]
\linespread{1.0}
\centering
\small
\begin{tabular}{p{2.6cm}p{1.9cm}<{\centering}p{1.9cm}<{\centering}}
\toprule
\textbf{Train on STAC}   & \textbf{Link} & \textbf{Link+Rel.} \\
\midrule
\multicolumn{3}{l}{Deep Sequential Parser \cite{shi2019deepSeqParser}} \\
Test on STAC & 73.1 & 55.7 \\
Test on Molweni & 58.6 & 26.2 \\
\midrule
\midrule
\multicolumn{3}{l}{Our Transformer-Based Parser} \\
Test on STAC & 73.4 & 55.5 \\
Test on Molweni & 57.8 & 26.4 \\
\midrule
\multicolumn{3}{l}{+ Utilizing Language Backbone} \\
Test on STAC & 75.3 [2.5\% $\uparrow$] & 56.9 [2.5\% $\uparrow$] \\
Test on Molweni & 60.7 [5.0\% $\uparrow$] & 31.5 [19.3\% $\uparrow$] \\
\midrule
\multicolumn{3}{l}{+ Cross-Domain Pre-training} \\
Test on STAC & 75.1 [2.3\% $\uparrow$] & 57.1 [2.8\% $\uparrow$] \\
Test on Molweni & 62.1 [7.4\% $\uparrow$] & 32.6 [23.4\% $\uparrow$] \\
\midrule
\multicolumn{3}{l}{+ Cross-Domain Vocabulary Refinement} \\
Test on STAC & 75.3 [2.3\% $\uparrow$] & 57.1 [2.8\% $\uparrow$] \\
Test on Molweni & 63.2 [9.3\% $\uparrow$] & 33.1 [25.3\% $\uparrow$]  \\
\bottomrule
\end{tabular}
\caption{\label{result-stac-table} Micro-F1 scores of link and relation prediction with models trained on \textbf{STAC}. Values in brackets denote relative increase over the base model.}
\vspace{-0.2cm}
\end{table}

\section{Experimental Result and Analysis}
\subsection{Configuration} 
The proposed models were implemented using PyTorch \citep{paszke2019pytorch} and Hugging Face\footnote{https://github.com/huggingface/transformers}. Learning rate was set at 2e-5, and the AdamW \cite{loshchilov2017-adamW} optimizer was applied. We trained each model for 20 epochs, and selected the best checkpoints based on evaluation scores. Input dialogue sequences were processed with the sub-word tokenization scheme used in \textit{`RoBERTa-base'} \cite{liu2019roberta}.

At the inference stage, we adopted the micro-averaged F1 score as the evaluation metric. Results of different settings are shown in Table \ref{result-join-table}-\ref{result-mol-table}. \textit{``Link''} denotes link prediction, and \textit{``Link+Rel.''} stands for a prediction that the dependency link and relation type are correct at the same time.

\subsection{Joint Domain Evaluation}
To compare performance between single-domain and joint-domain training, we obtain the upper bound parsing results on the merged data of two dialogue discourse treebanks (STAC and Molweni). As shown in Table \ref{result-join-table}, models trained on merged data achieve favorable results on both corpora, and perform slightly better than single-domain training. Moreover, our Transformer-based model with the language backbone outperforms the previous state-of-the-art baseline.

\begin{table}[t!]
\linespread{1.0}
\centering
\small
\begin{tabular}{p{2.6cm}p{1.9cm}<{\centering}p{1.9cm}<{\centering}}
\toprule
\textbf{Train on Molweni}  & \textbf{Link} & \textbf{Link+Rel.} \\
\midrule
\multicolumn{3}{l}{Deep Sequential Parser \cite{shi2019deepSeqParser}} \\
Test on STAC & 42.5 & 18.3 \\
Test on Molweni & 77.9 & 54.4 \\
\midrule
\midrule
\multicolumn{3}{l}{Our Transformer-Based Parser} \\
Test on STAC & 42.3 & 18.0 \\
Test on Molweni & 75.9 & 52.5 \\
\midrule
\multicolumn{3}{l}{+ Utilizing Language Backbone} \\
Test on STAC & 48.3 [14.2\% $\uparrow$] & 26.6 [47.7\% $\uparrow$] \\
Test on Molweni & 79.7 [5.1\% $\uparrow$] & 55.9 [6.4\% $\uparrow$] \\
\midrule
\multicolumn{3}{l}{+ Cross-Domain Pre-training} \\
Test on STAC & 48.8 [15.3\% $\uparrow$] & 28.4 [57.7\% $\uparrow$] \\
Test on Molweni & 79.6 [5.0\% $\uparrow$] & 55.7 [6.1\% $\uparrow$] \\
\midrule
\multicolumn{3}{l}{+ Cross-Domain Vocabulary Refinement} \\
Test on STAC & 50.5 [19.4\% $\uparrow$] & 28.9 [60.6\% $\uparrow$] \\
Test on Molweni & 79.5 [5.0\% $\uparrow$] &  55.7 [6.1\% $\uparrow$] \\
\bottomrule
\end{tabular}
\caption{\label{result-mol-table} Micro-F1 scores of link and relation prediction with models trained on \textbf{Molweni}. Values in brackets denote relative increase over the base model.}
\vspace{-0.2cm}
\end{table}

\subsection{Cross-Domain Evaluation}
To evaluate the effectiveness of the proposed domain integration methods, we conduct single-corpus training and cross-corpus evaluation (each treebank represents one dialogue domain).

For single-corpus training on STAC, as shown in Table \ref{result-stac-table}, the cross-domain performance on Molweni data of all models drops significantly, especially the relation prediction. Utilizing language backbone brings substantial improvement. This shows that linguistic features can be shared by samples from different treebanks under the SDRT theory. Adopting cross-domain pre-training and vocabulary refinement further improve the performance, and do not affect the original domain. Combining three methods provides the parser a relative 25.3\% improvement on the link+relation F1.

For single-corpus training on Molweni, as shown in Table \ref{result-mol-table}, baseline models obtain low link+relation F1 scores (around 18.0) on the STAC corpus. 
Noteworthy, the performance decrease of  \textit{STAC(train)}->\textit{Molweni(test)} is smaller than that of \textit{Molweni(train)}->\textit{STAC(test)}, we speculate that this may stem from a larger linguistic diversity in STAC data.
The scores are significantly elevated by adopting language backbone, cross-domain pre-training, and vocabulary refinement, achieving a relative 60.6\% improvement on link+relation F1.

\section{Conclusion}
In this paper, we investigated the domain-level generality of dialogue discourse parsing. Since existing corpora are collected from different conversation scenarios, models with single-domain training cannot perform well in other domains. The statistical analysis and experimental results suggest that domain adaptation or integration is necessary when neural parsers are applied in practical use cases, and utilizing prior language knowledge and adopting cross-domain pre-training can improve their generality.

\section*{Acknowledgments}
This research was supported by funding from the Institute for Infocomm Research (I2R) under A*STAR ARES, Singapore. We thank Ai Ti Aw for the insightful discussions. We also thank the anonymous reviewers for their precious feedback to help improve and extend this piece of work.

\bibliography{acl}
\bibliographystyle{acl_natbib}

\newpage

\appendix


\end{document}